%% file: HyperReservoir_arXiv.tex
\documentclass[
    amsmath,
    amssymb,
    reprint
]{revtex4-2}

\usepackage{graphicx}
\usepackage{bm}
\usepackage{dcolumn}
\usepackage{amsmath}
\usepackage{amssymb}

\input{macros}

\begin{document}

\title[Context-dependent time-series prediction via HyperReservoirs]
{Context-dependent time-series prediction via HyperReservoirs}

%
\author{Kohei Tsuchiyama}

\author{Takatomo Mihana}
 
\author{Ryoichi Horisaki}

\author{Andr\'{e} R\"{o}hm}%
\email{roehm@g.ecc.u-tokyo.ac.jp}
 
\affiliation{%
Department of Information Physics and Computing, Graduate School of Information Science and Technology,
The University of Tokyo, 7-3-1 Hongo, Bunkyo-ku, Tokyo 113-8656, Japan
}%

\date{September 2026}


\begin{abstract}
Time series prediction is a common application of reservoir computing.
When the training and testing time series data contains multiple dynamical regimes, because an underlying parameter is changing, or the data in fact consists of multiple distinct systems, simple application of the reservoir computing principle produces high prediction errors.
Here, we propose a HyperReservoir as an extended model of reservoir computing especially designed for such cases.
The HyperReservoir combines a main reservoir with a smaller context reservoir, where the latter modulates the output weights of the former. 
This structure resembles the hypernetworks from deep neural network literature.
However, in contrast, HyperReservoirs retain the simple training via linear regression of standard reservoir computing.
We compare the proposed architecture with a conventional ESN, in which context acts at the input, and a full-matrix Conceptor, in which context modulates the reservoir state space.
We evaluate all three models on time-series prediction tasks based on Lorenz and R\"ossler systems, including for varying bifurcation parameters and time sampling scales.
We find that the HyperReservoir achieves the lowest mean test error in all three tasks, and particularly outperforms conceptors on data that is sampled from the same attractor but at different time scales.
\end{abstract}

\maketitle

\input{sections/01_introduction}
\input{sections/02_contextual_reservoir}
\input{sections/03_numerical_experiments}
\input{sections/04_multifunctional_prediction}
\input{sections/05_shared_attractor}
\input{sections/06_contextual_readout}
\input{sections/07_conclusion}

\section*{SUPPLEMENTARY MATERIAL}
Supplemental Material is included after the main text and provides numerical settings and data-generation details, seedwise results, context-dimension and readout comparisons, same-attractor analyses, readout and storage counts, and software environment details.

\begin{acknowledgments}
This study was supported in part by a Grant-in-Aid for Transformative Research Areas (A) (JP22H05197), a Grant-in-Aid for JSPS Fellows (JP24KJ0868), Grant-in-Aid for Exploratory Research (JP25K22227),  JST-ALCA-Next (JPMJAN25F1), JST FOREST Program (JPMJFR2448) and SECOM Science and Technology Foundation.
\end{acknowledgments}

\vspace{5mm}

\section*{AUTHOR DECLARATIONS}

\noindent {\bf Conflicts of Interest}

The authors have no conflicts to disclose.

\vspace{5mm}

\noindent {\bf Author Contributions}
\noindent {\bf Kohei Tsuchiyama:} Conceptualization (lead); Methodology (lead); Formal analysis (lead); Investigation (lead); Visualization (lead); Writing - original draft (lead); Writing - review \& editing (equal).
{\bf Takatomo Mihana:} Investigation (supporting); Writing - review \& editing (supporting). 
{\bf Ryoichi Horisaki:} Writing - review \& editing (supporting).
{\bf Andr\'{e} R\"{o}hm:} Conceptualization (supporting); Methodology (supporting); Formal analysis (supporting); Investigation (supporting); Writing - original draft (supporting); Writing - review \& editing (equal).

\section*{Data Availability}
The code, numerical data, and figure-generation scripts that support the findings of this study are available from the corresponding author upon reasonable request.

\bibliography{reference}


\end{document}


\title{
Supplemental Material for \\
Context-dependent time-series prediction via HyperReservoirs
}


\author{Kohei Tsuchiyama}
\author{Takatomo Mihana}
\author{Ryoichi Horisaki}
\author{Andr\'{e} R\"{o}hm}

\date{\today}

\maketitle

\section{Numerical settings and data generation}
\label{sec:supp_numerical_details}
This section gives the numerical settings used in all experiments.
The same generated datasets are used for all architectures, while the random recurrent realization is varied independently over three model seeds.
Table~\ref{tab:supp_common_settings} summarizes the common settings.

\begin{table*}[t]
\caption{
Common numerical and model settings used in the three contextual prediction tasks.
The model seed changes the random recurrent realization while the generated datasets remain fixed.
}
\label{tab:supp_common_settings}
\centering
\begin{tabular}{lll}
\toprule
Category & Quantity & Value \\
\midrule
Data
& data seed
& $0$
\\
&
model seeds
& $\{0,1,2\}$
\\
&
training trajectories
& $\NumTrainTrajectories$
\\
&
validation trajectories
& $\NumValTrajectories$
\\
&
test trajectories
& $\NumTestTrajectories$
\\
&
samples per trajectory
& $\TrajectoryLength$
\\
&
prediction horizon
& $H=\PredictionHorizon$
\\
&
reservoir washout
& $\WashoutLength$ samples
\\[2pt]

Dynamical-system integration
& integrator
& fourth-order Runge--Kutta
\\
&
internal integration step
& $0.005$
\\
&
sampling interval
& $0.05$
\\
&
dynamical burn-in
& $2000$ internal steps
\\[2pt]

Reservoir
& total recurrent-state budget
& $N_{\mathrm{tot}}=\TotalStateBudget$
\\
&
main-reservoir spectral radius
& $0.9$
\\
&
context-reservoir spectral radius
& $0.9$
\\
&
main-reservoir input scale
& $0.3$
\\
&
context-reservoir input scale
& $0.1$
\\
&
ESN/Conceptor leak coefficient
& $\alpha = 0.1$
\\
&
HyperReservoir main leak coefficient
& $\alpha^\mathrm{R} = 0.1$
\\
&
HyperReservoir context leak coefficient
& $\alpha^\mathrm{H} = 1/60$
\\
&
nonlinearity
& $\tanh$
\\[2pt]

Readout
& ridge coefficient
& $\lambda=\RidgeAlpha$
\\
&
output bias
& fitted and unregularized
\\[2pt]

HyperReservoir selection
& context dimensions
& $M\in\{2,5,10,20,40\}$
\\
&
main dimension
& $N=120-M$
\\[2pt]

Conceptor selection
& initial aperture candidates
& $\gamma\in\{1,3,10,30\}$
\\
&
additional Lorenz--Rössler refinement
& $\gamma\in\{1.5,2,4,5,7\}$
\\
\bottomrule
\end{tabular}
\end{table*}

\subsection{Trajectory generation}
\label{subsec:supp_trajectory_generation}
All continuous-time systems are integrated with a fourth-order Runge--Kutta scheme using an internal integration step of $0.005$.
The states are recorded every ten integration steps, giving a sampling interval of $0.05$.
Each recorded trajectory contains $1000$ samples.

Before recording, each system is integrated for $2000$ internal steps to reduce dependence on the initial condition sampled.
For the R\"ossler systems, the initial conditions are drawn independently as
\begin{equation}
    x_0 \sim \mathcal{N}(0,1),
    \qquad
    y_0 \sim \mathcal{N}(0,1),
    \qquad
    z_0 \sim \mathcal{N}(0.5,0.2^2).
    \label{eq:supp_rossler_initial_conditions}
\end{equation}
For the Lorenz system,
\begin{equation}
    x_0 \sim \mathcal{N}(0,1),
    \qquad
    y_0 \sim \mathcal{N}(0,1),
    \qquad
    z_0 \sim \mathcal{N}(20,2^2).
    \label{eq:supp_lorenz_initial_conditions}
\end{equation}
No observation noise or process noise is added.
As a numerical safeguard, each dynamical-state component is trimmed to $[-80,80]$ during trajectory generation.

The training, validation and test sets contain $\NumTrainTrajectories$, $\NumValTrajectories$, and $\NumTestTrajectories$ independently initialized trajectories, respectively.
The same generated datasets are used for all architectures.
Each architecture is evaluated using three independently initialized reservoir realizations.

\subsection{Reservoir initialization}
\label{subsec:supp_reservoir_initialization}
For a reservoir of dimension $N$, the recurrent matrix is initialized with independent Gaussian entries of standard deviation $N^{-1/2}$ and then rescaled to spectral radius $0.9$.
The context-reservoir matrix is initialized independently using the same procedure.

Input weights are initialized independently from a zero-mean Gaussian distribution with standard deviation $\sigma_{\mathrm{in}}/\sqrt{U}$, where $U$ is the total number of available observation and context channels.
Input channels not used by a particular reservoir are then masked out according to the routing summarized in Table~\ref{tab:supp_context_routing}.
The input scales are
\begin{equation}
    \sigma_{\mathrm{in}}^{\mathrm{R}}=0.3,
    \qquad
    \sigma_{\mathrm{in}}^{\mathrm{H}}=0.1.
    \label{eq:supp_input_scales}
\end{equation}
The recurrent and input matrices are initialized as described above, and
the bias terms in both reservoir-state updates are set to zero.

The leak coefficient is $\alpha=0.1$ for the context-input ESN and Conceptor reservoir. 
For the HyperReservoir, the main and context reservoirs use
\begin{equation}
    \alpha^{\mathrm{R}}=0.1,
    \qquad
    \alpha^{\mathrm{H}}=\frac{1}{60},
\end{equation}
respectively.

\subsection{Context signals}
\label{subsec:supp_context_routing}
For the Lorenz--R\"ossler task, the context is a two-dimensional one-hot vector identifying the active system.
For the related-R\"ossler task, a two-dimensional one-hot vector identifies $c_{\mathrm{R}}=3.5$ or $5.7$.
For the same-attractor task,
\begin{equation}
    \mathbf{c}
    =
    \begin{bmatrix}
        c_{\mathrm{R}} & \nu
    \end{bmatrix}^{\mathsf T},
    \qquad
    c_{\mathrm{R}}=5.7,
    \qquad
    \nu\in\{0.5,1.5\}.
\end{equation}
In the HyperReservoir, these contextual channels are supplied only to the context reservoir.

\begin{table*}[t]
\caption{
Architecture-specific routing of the observed dynamical state and context.
}
\label{tab:supp_context_routing}
\centering
\begin{tabular}{llll}
\toprule
Model
& Main reservoir
& Context reservoir / selector
& Context-dependent operation
\\
\midrule
Context-input ESN
& observation + context
& none
& input forcing
\\
full-matrix Conceptor
& observation only
& context selects $C_k$
& state-space filtering
\\
HyperReservoir
& observation only
& context only
& readout modulation
\\
\bottomrule
\end{tabular}
\end{table*}

\subsection{Normalization}
\label{subsec:supp_normalization}
For all three tasks, normalization parameters are estimated exclusively from the training split and are subsequently applied without modification to the validation and test sets.
For component $d$,
\begin{equation}
    \widetilde{u}_{t,d}
    =
    \frac{
        u_{t,d}-\mu^{\mathrm{train}}_{u,d}
    }{
        \sigma^{\mathrm{train}}_{u,d}
    },
    \qquad
    \widetilde{y}_{t,d}
    =
    \frac{
        y_{t,d}-\mu^{\mathrm{train}}_{y,d}
    }{
        \sigma^{\mathrm{train}}_{y,d}
    }.
    \label{eq:supp_normalization}
\end{equation}
The standard deviations are lower bounded by $10^{-6}$.
The same transformation is reused for validation and test data without re-estimating any statistics.

For the Lorenz--R\"ossler task, the two dynamical systems are not standardized separately.
Instead, a single set of normalization statistics is estimated from the combined training trajectories and applied to both contextual regimes.
This avoids introducing system identity through context-dependent preprocessing.

\subsection{Readout fitting}
\label{subsec:supp_readout_fitting}
All output coefficients are fitted by ridge regression after the washout interval.
Let $X$ denote the matrix of valid readout features and $Y$ the corresponding targets.
An unregularized constant column is appended to the feature matrix:
\begin{equation}
    X_{\mathrm{aug}}
    =
    \begin{bmatrix}
        X & \mathbf{1}
    \end{bmatrix}.
    \label{eq:supp_augmented_design_matrix}
\end{equation}
The fitted coefficients are
\begin{equation}
    W_{\mathrm{aug}}
    =
    \left(
        X_{\mathrm{aug}}^{\mathsf T}
        X_{\mathrm{aug}}
        +
        \lambda R
    \right)^{-1}
    X_{\mathrm{aug}}^{\mathsf T}Y.
    \label{eq:supp_ridge_solution}
\end{equation}
Here, $R$ is the identity matrix except that the entry corresponding to the constant feature is zero.
Thus, the output bias is not regularized.
The ridge coefficient is fixed to $\lambda=10^{-4}$ in all principal experiments and is not selected from validation data.

\subsection{Conceptor estimation}
\label{subsec:supp_conceptor_estimation}

Conceptors are estimated exclusively from post-washout training reservoir states.
For context $k$, the uncentered empirical correlation matrix is
\begin{equation}
    R_k
    =
    \frac{1}{n_k}
    \sum_{i\in\mathcal{I}_k}
    \mathbf{x}_i\mathbf{x}_i^{\mathsf T},
\end{equation}
where $\mathcal{I}_k$ indexes training states in context $k$.
The corresponding full-matrix Conceptor is
\begin{equation}
    C_k
    =
    R_k
    \left(
        R_k+\gamma^{-2}I
    \right)^{-1}.
\end{equation}
No validation or test states are used in estimating $R_k$ or $C_k$.
The readout is then trained from the Conceptor-filtered training states.

\subsection{Validation-based architecture selection}
\label{subsec:supp_validation_selection}
The HyperReservoir context dimension is evaluated over
\begin{equation}
    M\in\{2,5,10,20,40\},
    \qquad
    N=120-M.
    \label{eq:supp_M_candidates}
\end{equation}
The full-matrix Conceptor aperture is initially evaluated over
\begin{equation}
    \gamma\in\{1,3,10,30\}.
    \label{eq:supp_aperture_candidates}
\end{equation}
For the Lorenz--R\"ossler task, the aperture search is additionally refined around the best-performing region of the initial validation sweep using
\begin{equation}
    \gamma\in\{1.5,2,4,5,7\}.
    \label{eq:supp_step1_aperture_refinement}
\end{equation}

For a candidate configuration $\theta$, let $E_{\mathrm{val}}^{(s)}(\theta)$ denote its validation cNMSE for reservoir seed $s$.
One task-level configuration is selected according to
\begin{equation}
    \theta^\star
    =
    \underset{\theta}{\operatorname{arg\,min}}
    \frac{1}{3}
    \sum_{s\in\{0,1,2\}}
    E_{\mathrm{val}}^{(s)}(\theta).
    \label{eq:supp_validation_selection_rule}
\end{equation}
All numerical architecture and hyperparameter selections use validation cNMSE; test cNMSE is not included in the numerical selection criterion.
A separate architecture is therefore not selected from the test performance of each seed.

For the augmented HyperReservoir, the validation-selected dimensions are
\begin{equation}
\begin{aligned}
    M^\star_{\mathrm{distinct}} &= 10,
    &\qquad
    N^\star_{\mathrm{distinct}} &= 110,
    \\
    M^\star_{\mathrm{related}} &= 5,
    &
    N^\star_{\mathrm{related}} &= 115,
    \\
    M^\star_{\mathrm{same}} &= 5,
    &
    N^\star_{\mathrm{same}} &= 115.
\end{aligned}
    \label{eq:supp_selected_M}
\end{equation}

For the full-matrix Conceptor model, the final task-level apertures selected by mean validation cNMSE across the three reservoir realizations are
\begin{equation}
\begin{aligned}
    \gamma^\star_{\mathrm{distinct}} &= 4,
    \\
    \gamma^\star_{\mathrm{related}} &= 1,
    \\
    \gamma^\star_{\mathrm{same}} &= 1.
\end{aligned}
    \label{eq:supp_selected_apertures}
\end{equation}

The same task-level aperture is used for every reservoir realization and in all Conceptor analyses associated with a given task, including the principal prediction comparison and the same-attractor operator analysis.

\subsection{Evaluation-mask details}
\label{subsec:supp_metric_details}
The first $\WashoutLength$ samples of each trajectory are excluded as reservoir washout.
The final $H$ samples are also excluded because they do not have a valid future-state target.
The component-normalized mean squared error is evaluated on the remaining samples using targets normalized with training-derived statistics.
The numerical stabilization constant in the denominator is $10^{-4}$.
No validation- or test-set recentering is performed.

The arithmetic mean and sample standard deviation across the three reservoir realizations are reported only as descriptive summaries.
Because the number of realizations is small, individual seed values are shown in the principal figures.

\section{Main-comparison results}
\label{sec:supp_main_results}

Table~\ref{tab:supp_main_seedwise} reports the individual test cNMSE values for all three random reservoir realizations after validation-based architecture selection.

\begin{table*}[t]
\caption{
Seedwise test cNMSE.
The final column gives the arithmetic mean and sample standard deviation over the three random reservoir realizations.
}
\label{tab:supp_main_seedwise}
\centering
\begin{tabular}{llcccc}
\toprule
Task
& Model
& Seed 0
& Seed 1
& Seed 2
& Mean $\pm$ sample SD
\\
\midrule

\multirow{3}{*}{Distinct}
& ESN
& $4.5365\times10^{-2}$
& $4.4232\times10^{-2}$
& $4.4425\times10^{-2}$
& $(4.4674\pm0.0606)\times10^{-2}$
\\
&
Conceptor
& $4.7693\times10^{-2}$
& $4.2918\times10^{-2}$
& $4.1809\times10^{-2}$
& $(4.4140\pm0.3126)\times10^{-2}$
\\
&
HyperReservoir
& $3.2950\times10^{-2}$
& $3.9176\times10^{-2}$
& $3.1146\times10^{-2}$
& $(3.4424\pm0.4213)\times10^{-2}$
\\[3pt]

\multirow{3}{*}{Related}
& ESN
& $4.6593\times10^{-4}$
& $5.7953\times10^{-4}$
& $3.4698\times10^{-4}$
& $(4.6415\pm1.1629)\times10^{-4}$
\\
&
Conceptor
& $4.1522\times10^{-3}$
& $3.5667\times10^{-3}$
& $8.5255\times10^{-3}$
& $(5.4148\pm2.7098)\times10^{-3}$
\\
&
HyperReservoir
& $8.7665\times10^{-5}$
& $7.3333\times10^{-5}$
& $7.5406\times10^{-5}$
& $(7.8801\pm0.7746)\times10^{-5}$
\\[3pt]

\multirow{3}{*}{Same attractor}
& ESN
& $4.9357\times10^{-3}$
& $5.4010\times10^{-3}$
& $3.8899\times10^{-3}$
& $(4.7422\pm0.7739)\times10^{-3}$
\\
&
Conceptor
& $6.7328\times10^{-3}$
& $7.6255\times10^{-3}$
& $1.7936\times10^{-2}$
& $(1.0765\pm0.6226)\times10^{-2}$
\\
&
HyperReservoir
& $6.6345\times10^{-4}$
& $5.3148\times10^{-4}$
& $7.3756\times10^{-4}$
& $(6.4416\pm1.0439)\times10^{-4}$
\\

\bottomrule
\end{tabular}
\end{table*}

\section{Context-reservoir dimension sweep}
\label{sec:supp_M_sweep}

The context dimension is varied while keeping the total recurrent-state budget fixed:
\begin{equation}
    N+M=120.
    \label{eq:supp_M_budget}
\end{equation}
The tested context dimensions are
\begin{equation}
    M\in\{2,5,10,20,40\}.
    \label{eq:supp_M_sweep_values}
\end{equation}

Table~\ref{tab:supp_M_sweep} gives the test summaries for all three tasks.
The architecture used in the comparison is selected from validation performance rather than from the test values in this table.
Panel~(c) of Fig.~5 in the main manuscript shows the related-R\"ossler sweep, for which the validation-selected value is $M^\star=5$.

\begin{table*}[t]
\caption{
Test cNMSE for the augmented HyperReservoir as the context-reservoir dimension $M$ is varied under the fixed recurrent-state budget $N+M=120$.
Values are the mean and sample standard deviation across three reservoir seeds.
}
\label{tab:supp_M_sweep}
\centering
\begin{tabular}{cccc}
\toprule
$M$
& Distinct
& Related
& Same attractor
\\
\midrule
2
&
$(3.9722\pm0.3106)\times10^{-2}$
&
$(1.8801\pm0.3013)\times10^{-4}$
&
$(1.6162\pm0.8627)\times10^{-3}$
\\

5
&
$(3.5676\pm0.1228)\times10^{-2}$
&
$(7.8801\pm0.7746)\times10^{-5}$
&
$(6.4416\pm1.0439)\times10^{-4}$
\\

10
&
$(3.4424\pm0.4213)\times10^{-2}$
&
$(1.1040\pm0.1318)\times10^{-4}$
&
$(6.6432\pm1.4084)\times10^{-4}$
\\

20
&
$(3.5997\pm0.2987)\times10^{-2}$
&
$(1.4243\pm0.1530)\times10^{-4}$
&
$(8.4349\pm3.8616)\times10^{-4}$
\\

40
&
$(3.7962\pm0.3321)\times10^{-2}$
&
$(2.5571\pm1.2052)\times10^{-4}$
&
$(7.7254\pm1.5993)\times10^{-4}$
\\
\bottomrule
\end{tabular}
\end{table*}

The validation-selection procedure chooses $M=10$ for the geometrically distinct Lorenz--R\"ossler task and $M=5$ for both the related-R\"ossler and same-attractor tasks.
Increasing $M$ does not produce a monotonic improvement in test prediction accuracy.
Under the matched-state constraint, increasing the number of context-reservoir states simultaneously reduces the dimension available to the main reservoir and changes the size of the bilinear readout.

\section{Readout-comparison results}
\label{sec:supp_readout_ablation}

Here, we present details for the three variations of the HyperReservoir: As explained in the main manuscript, there are several ways for combining the features of the main and context reservoir. 
Here, we chose the following:
\begin{equation}
\begin{aligned}
    \boldsymbol{\psi}^{\mathrm{concat}}_t
    &=
    \begin{bmatrix}
        \mathbf{h}^{\mathrm{R}}_t \\
        \mathbf{h}^{\mathrm{H}}_t
    \end{bmatrix},
    \\
    \boldsymbol{\psi}^{\mathrm{strict}}_t
    &=
    \mathbf{h}^{\mathrm{H}}_t
    \otimes
    \mathbf{h}^{\mathrm{R}}_t,
    \\
    \boldsymbol{\psi}^{\mathrm{aug}}_t
    &=
    \begin{bmatrix}
        \mathbf{h}^{\mathrm{R}}_t \\
        \mathbf{h}^{\mathrm{H}}_t \\
        \mathbf{h}^{\mathrm{H}}_t
        \otimes
        \mathbf{h}^{\mathrm{R}}_t
    \end{bmatrix}.
\end{aligned}
\label{eq:supp_readout_features}
\end{equation}

Table~\ref{tab:supp_readout_ablation} gives the corresponding test performance.

\begin{table*}[t]
\caption{
Readout comparison using the validation-selected $N$ and $M$ of the augmented HyperReservoir for each task.
Values are mean test cNMSE and sample standard deviation across three random reservoir realizations.
}
\label{tab:supp_readout_ablation}
\centering
\begin{tabular}{lccc}
\toprule
Readout
& Distinct
& Related
& Same attractor
\\
\midrule
Concat
&
$(4.4320\pm0.3341)\times10^{-2}$
&
$(2.0709\pm0.3444)\times10^{-4}$
&
$(2.0795\pm0.1217)\times10^{-3}$
\\
Strict multiplicative
&
$(5.3592\pm0.6946)\times10^{-2}$
&
$(9.1011\pm2.2018)\times10^{-2}$
&
$(1.9783\pm0.9186)\times10^{-3}$
\\
Augmented
&
$(3.4424\pm0.4213)\times10^{-2}$
&
$(7.8801\pm0.7746)\times10^{-5}$
&
$(6.4416\pm1.0439)\times10^{-4}$
\\
\bottomrule
\end{tabular}
\end{table*}

The Augmented construction gives the lowest mean cNMSE in all three settings.
The Strict construction performs particularly poorly for the related-R\"ossler setting.
All three readouts use identical reservoir realizations, data partitions, and fitting conventions within each setting.




\section{Detailed same-attractor analyses}
\label{sec:supp_same_attractor}

Panels 4(a) and 4(b) of the main manuscript show one-step prediction errors. 
For the displayed $\chi$ component, the residual is defined as
\begin{align*}
e_{\chi,t} = \hat{\chi}_{t+1} - \chi_{t+1}.
\end{align*}
The raw prediction traces of the three models closely overlap on the state scale, so the residual representation is used to expose local differences among the architectures without changing the underlying prediction task or evaluation data.

The representative reservoir realization is chosen without using test performance. 
Among the reservoir seeds common to the conventional ESN, full-matrix Conceptor, and HyperReservoir, seeds are ranked separately by validation cNMSE for each architecture, and the seed with the lowest mean validation rank across the three models is used for visualization.
The same selected seed is used for all models.

The slow and fast panels use the same fixed display window of 80 samples, beginning at sample 150 of the selected test trajectories. 
With the sampling interval of $0.05$ and prediction horizon $H=1$, the displayed future-target times are approximately $7.55–11.50$. 
The same window is used for all models. 
The vertical axes of the slow and fast panels are scaled independently and symmetrically around zero because the absolute error magnitudes differ substantially between the two temporal regimes. 
This scaling affects only the visualization; all quantitative comparisons in the main text use cNMSE evaluated over the valid test set.

\subsection{Wrong-context intervention}
\label{subsec:supp_wrong_context}

To directly test whether each architecture functionally uses the supplied context, we perform a counterfactual context intervention in the same-attractor temporal-scale task.
For true context $i$ and supplied context $j$, we define
\begin{equation}
    E_{ij}
    =
    \mathrm{cNMSE}
    \left(
        \text{true context }i,\,
        \text{supplied context }j
    \right).
    \label{eq:supp_wrong_context_error}
\end{equation}
The diagonal entries correspond to prediction with the correct context, whereas the off-diagonal entries correspond to prediction after replacing the supplied context while keeping the observed trajectory unchanged.

The summary quantity reported in Fig.~4(c) of the main manuscript is
\begin{equation}
    R_{\mathrm{ctx}}
    =
    \frac{E_{11}+E_{22}}
         {E_{12}+E_{21}},
    \label{eq:supp_context_ratio}
\end{equation}
which is equivalently the mean correct-context cNMSE divided by the mean wrong-context cNMSE.
A value substantially below unity therefore indicates that replacing the supplied context strongly degrades prediction accuracy.

The intervention is implemented according to the context-dependent pathway of each architecture.
For the context-input ESN, the context channels in the reservoir input are replaced.
For the full-matrix Conceptor, the context selects the corresponding context-specific Conceptor matrix.
For the HyperReservoir, the explicit context supplied to the context reservoir is replaced while the main reservoir continues to receive the same observed trajectory.
For every true/supplied-context pair, the complete input sequence is passed through a new model forward pass.
Thus, whenever context affects a recurrent state, that state is recomputed under the counterfactual context rather than reused from the correct-context trajectory.

\begin{figure*}[t]
    \centering
    \includegraphics[width=0.9\textwidth]
        {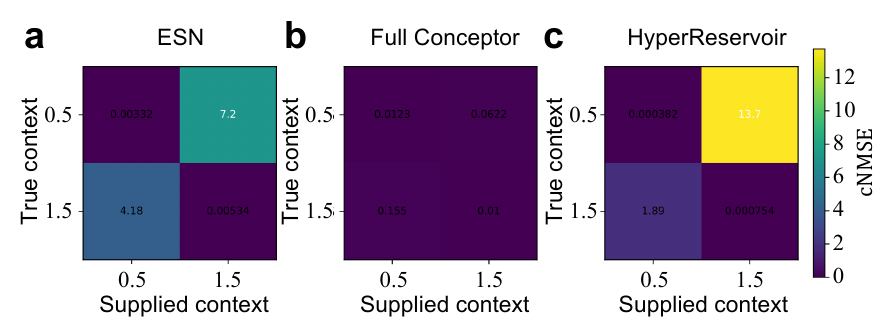}
    \caption{
    Counterfactual context dependence in the same-attractor temporal-scale task.
    Mean cNMSE matrices for
    (a) the context-input ESN,
    (b) the full-matrix Conceptor, and
    (c) the HyperReservoir.
    Rows indicate the true temporal-scale context and columns indicate the supplied context, with $\nu\in\{0.5,1.5\}$.
    Diagonal entries therefore correspond to the correct context, whereas off-diagonal entries correspond to counterfactual context replacement.
    Each matrix is averaged over the three reservoir realizations after task-level validation-based model selection.
    A common color scale is used across the three architectures.
    }
    \label{fig:supp_wrong_context}
\end{figure*}

Figure~\ref{fig:supp_wrong_context} shows the individual error terms underlying the summary ratio $R_{\mathrm{ctx}}$ in Fig.~4(c) of the main manuscript.
For both the conventional ESN and the HyperReservoir, replacing the supplied context increases the prediction error by several orders of magnitude relative to the corresponding diagonal entries.
This separation is especially pronounced for the HyperReservoir, which combines low correct-context errors with large errors under counterfactual context replacement.
By contrast, the full-matrix Conceptor exhibits a substantially smaller separation between its diagonal and off-diagonal errors.
These matrices therefore provide the direct error-level view underlying the smaller $R_{\mathrm{ctx}}$ values of the ESN and, most strongly, the HyperReservoir.

\subsection{Full-matrix Conceptor comparison}
\label{subsec:supp_conceptor_similarity}

\begin{figure*}[t]
    \centering
    \includegraphics[width=0.6\textwidth]{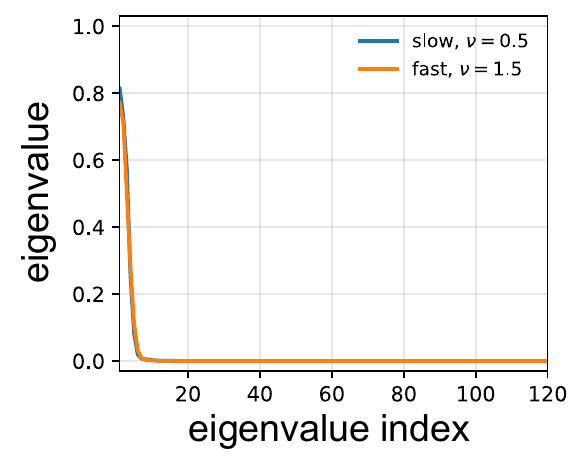}
    \caption{
    Ordered eigenvalue spectra of the slow and fast Conceptors in the same-attractor temporal-scale task.
    The curves show the spectra of $C_{\mathrm{slow}}$ and $C_{\mathrm{fast}}$ for the representative reservoir realization (seed 1), using the task-level validation-selected aperture $\gamma^\ast=1$.
    Both Conceptors are estimated from post-washout training reservoir states.
    This comparison is descriptive only: similarity of the eigenvalue spectra does not by itself imply similarity of the corresponding operators, because their eigenvectors are not considered.
    }
    \label{fig:eigenvalue}
\end{figure*}

Let $C_{\mathrm{slow}}$ and $C_{\mathrm{fast}}$ denote the full-matrix
Conceptors estimated from training trajectories for the two temporal
regimes using the task-level validation-selected aperture $\gamma^\ast=1$.
In the main manuscript, we report the Frobenius cosine similarity between
these operators. As an additional descriptive comparison, Fig.~\ref{fig:eigenvalue} shows
their ordered eigenvalue spectra for the representative reservoir
realization. The spectra largely overlap; however, eigenvalue similarity
alone does not establish operator similarity because the corresponding
eigenvectors are not considered.

To examine whether the two Conceptors also differ in their spectral filtering profiles, we now also compare their ordered eigenvalue spectra. 
The ordered eigenvalue spectra shown in Fig.~\ref{fig:eigenvalue} largely overlap across the two regimes, consistent with the high Frobenius cosine similarity reported in Fig.~4(d) of the main manuscript.

\section{Fitted readout dimensions and stored contextual quantities}
\label{sec:supp_parameter_counts}

The matched recurrent-state budget controls the dimension of the dynamical state but does not equalize the number of fitted output coefficients or stored context-specific quantities.
This section gives the corresponding counts explicitly.

For $D=3$ outputs and $F$ nonconstant readout features,
\begin{align}
P_{\mathrm{out}}&=D(F+1)\\
&=3(F+1)
\end{align}
The conventional ESN and the shared Conceptor readout each have $F=120$, giving 363 fitted coefficients.
In addition, the two-context Conceptor model stores two $120\times120$ matrices, corresponding to 28,800 context-specific matrix entries.
Table~\ref{tab:supp_parameter_counts} gives the readout dimensions used in the HyperReservoir comparisons.

\begin{table}[t]
\caption{
Readout feature dimensions and fitted output coefficients for the recurrent-state allocations used in the readout ablation.
The Conceptor storage entry is not included because it is separate from the shared linear readout.
}
\label{tab:supp_parameter_counts}
\centering
\begin{tabular}{llrr}
\toprule
Allocation
& Model
& \makecell[r]{Feature\\dimension}
& \makecell[r]{Fitted readout\\coefficients}
\\
\midrule
\multirow{3}{*}{$N=110,\ M=10$}
& Concat & 120 & 363 \\
& Strict & 1100 & 3303 \\
& Augmented & 1220 & 3663 \\
\midrule
\multirow{3}{*}{$N=115,\ M=5$}
& Concat & 120 & 363 \\
& Strict & 575 & 1728 \\
& Augmented & 695 & 2088 \\
\bottomrule
\end{tabular}
\end{table}

\subsection{Dependence on context dimension}
\label{subsec:supp_M_parameter_counts}

Under the fixed state budget $N+M=120$, the augmented feature dimension is
\begin{equation}
    F_{\mathrm{aug}}(M)
    =
    120+M(120-M).
    \label{eq:supp_feature_dimension_M}
\end{equation}
The fitted readout count is therefore
\begin{equation}
    P_{\mathrm{aug}}(M)
    =
    3
    \left[
        121+M(120-M)
    \right].
    \label{eq:supp_parameter_dimension_M}
\end{equation}

\begin{table}[t]
\caption{
HyperReservoir feature dimension and fitted readout size across the context-dimension sweep under $N+M=120$.
}
\label{tab:supp_M_cost}
\centering
\begin{tabular}{rrrr}
\toprule
$M$
& $N$
& $F_{\mathrm{aug}}$
& $P_{\mathrm{aug}}$
\\
\midrule
2  & 118 & 356  & 1071 \\
5  & 115 & 695  & 2088 \\
10 & 110 & 1220 & 3663 \\
20 & 100 & 2120 & 6363 \\
40 & 80  & 3320 & 9963 \\
\bottomrule
\end{tabular}
\end{table}

Increasing $M$ therefore has two distinct effects under the matched-state constraint.
It reallocates recurrent states from the main reservoir to the context reservoir while simultaneously increasing the bilinear readout dimension over the evaluated range.
The $M$ sweep is thus not a simple increase in model size.

The conventional ESN readout scales as $\mathcal{O}(DN_{\mathrm{tot}})$.
A full-matrix Conceptor additionally requires a dense state-space multiplication of order $\mathcal{O}(N_{\mathrm{tot}}^2)$ when the filter is applied.
The augmented HyperReservoir constructs an $NM$ bilinear feature block and therefore incurs output-stage work that scales with $NM$ in addition to the linear readout.

\section{Software environment}
\label{sec:supp_reproducibility}

All final experiments were executed using Python 3.10.14 and PyTorch 2.7.1 on macOS 14.4 (arm64) with an Apple M3 processor (8 CPU cores, 16 GB memory).

\bibliography{}

%% file: macros.tex
\usepackage{xcolor}

%% file: sections/01_introduction.tex
\section{Introduction}
\label{sec:introduction}
High-dimensional recurrent dynamics provide a useful basis for representing and transforming temporal signals.
Reservoir computing makes use of this principle by retaining a high-dimensional dynamical system as a fixed nonlinear substrate while training only a comparatively simple output transformation~\cite{Jaeger2004,Maass2002}.
This restriction reduces the need for recurrent optimization and is particularly attractive when the underlying dynamics are readily available but difficult or costly to modify, as in many physical implementations~\cite{appeltant2011information,vandoorne2014experimental,Nakajima2015,kanao2019reservoir}.
However, a largely fixed recurrent substrate also limits how much the computation can be adapted to changing conditions.

Many computational systems must operate across multiple tasks, environmental conditions, or dynamical regimes rather than perform a single fixed computation~\cite{yang2019task,driscoll2024flexible,flynn2021multifunctionality}.
Reusing a common recurrent substrate across such conditions can allow a single recurrent system to support multiple dynamical or computational regimes~\cite{flynn2021multifunctionality,du2025multifunctional}. 
Flexible behavior in biological and artificial recurrent networks likewise relies on shared neural dynamics being recruited differently depending on task or context~\cite{mante2013context,yang2019task,driscoll2024flexible}. 
Importantly, similar observations or internal representations can require different responses under different contexts~\cite{mante2013context}. 
Therefore, contextual information is not merely an additional input variable; it specifies how a shared computational substrate should be used for the currently relevant mapping.

A natural question is then how contextual flexibility can be introduced without discarding recurrent structure that remains useful across regimes.
In particular, when a common recurrent representation remains informative across contexts, it may be advantageous to preserve that representation and adapt only the mapping from the representation to the required output.
This suggests a distinction between context-dependent modification of the recurrent representation itself and context-dependent interpretation of an otherwise shared representation.
Context can in principle act at several stages of a recurrent computation.
It may be supplied as an external input, as in parameter-aware or multifunctional reservoir computing~\cite{kong2021critical,xiao2021amplitude,du2025multifunctional}; it may restrict the accessible region of reservoir state space, as in Conceptors~\cite{jaeger2017using}; or it may alter how a shared reservoir representation is mapped to the output, as in reservoir architectures with multiple readouts~\cite{laan2017multiple,tanaka2023multiple}.
Related work has also extended reservoir readouts through nonlinear combinations of reservoir variables~\cite{ohkubo2024generalized}.
These alternatives impose different structural assumptions because they determine which parts of the computation remain shared across dynamical regimes and which are allowed to change with context.
The central issue is therefore not simply whether contextual information is available, but which components of the recurrent computation should remain common and where context-dependent degrees of freedom should be introduced.

This distinction becomes especially important when the target regimes cannot be readily distinguished from state-space geometry alone.
If a context change generates geometrically distinct attractors, the occupied regions of state space already provide substantial information about the active regime~\cite{jaeger2017using,Lu2018}.
This becomes more difficult when attractors from different contexts are similar, and is particularly restrictive when they share the same underlying attractor geometry.
A key example is temporal scaling.
Temporal scaling has been studied in recurrent neural dynamics, where similar or approximately invariant trajectories can be traversed at different speeds~\cite{wang2018flexible,goudar2018encoding}.
For example, consider two systems that differ only by a temporal-scale factor,
\begin{equation}
    \dot{\mathbf{s}} = \nu_{c}\mathbf{f}(\mathbf{s}),
    \label{eq:intro_temporal_scaling}
\end{equation}
where $\nu_{c}$ depends on context.
Changing $\nu_{c}$ changes the rate at which a trajectory advances through the attractor without changing the underlying continuous-time orbit geometry.
Consequently, the principal contextual distinction is not which region of state space is occupied, but how rapidly the state evolves through that region.
For future-state prediction, similar current states can therefore require different outputs depending on the temporal context.
In such a setting, the challenge is not necessarily to construct a different recurrent representation for every context, but to map a substantially shared representation to different context-dependent targets.

Motivated by this distinction, we propose the HyperReservoir, a reservoir-computing architecture for context-dependent decoding of shared recurrent dynamics.
The architecture retains a common main reservoir that represents the observed dynamics, while a smaller context reservoir encodes the supplied context and modulates the readout applied to the main-reservoir features.
Separate reservoir pathways for measurement and regime-parameter sequences have previously been used for multi-regime time-series prediction~\cite{zhong2017double}.
Here, the two representations are instead coupled through a bilinear readout, so that the context-reservoir state parameterizes the effective mapping from shared main-reservoir features to the output.
This interpretation is closely related to Hypernetworks, which provide a general framework for making selected parameters of a computation depend on contextual information~\cite{ha2017hypernetworks}.
Importantly, both recurrent subsystems remain fixed, and the complete readout is fitted by linear regression.

To evaluate the proposed readout-level modulation, we compare the HyperReservoir with two established alternatives that introduce the same contextual information at different stages of reservoir computation: a conventional context-input ESN, in which context enters through the reservoir input~\cite{jaeger2001echo,du2025multifunctional}, and full-matrix Conceptors, in which context acts through state-space restriction~\cite{jaeger2017using}.
Together, the three architectures provide input-, state-, and readout-level forms of contextual adaptation.

We evaluate these architectures on three future-state prediction settings: geometrically distinct Lorenz and R\"ossler systems, related R\"ossler regimes, and the same R\"ossler attractor traversed at different temporal scales.
These settings range from clearly distinct state-space geometries to a case in which the underlying continuous-time attractor geometry is shared while the required finite-time prediction remains context dependent.

Across all three settings, the augmented HyperReservoir achieves lower mean prediction error than the conventional context-input ESN, with larger differences for the related and shared-attractor regimes.
Additional analyses examine the functional use of the supplied context and the contribution of the context-dependent readout structure.
Together, the results support context-dependent decoding as a useful mechanism when recurrent features can remain substantially shared across regimes while the required predictive mapping changes with context.

%% file: sections/02_contextual_reservoir.tex
\section{Contextual adaptation in reservoir computing}
\label{sec:contextual_reservoir}

\begin{figure*}[t]
    \centering
    \includegraphics[width=\textwidth]{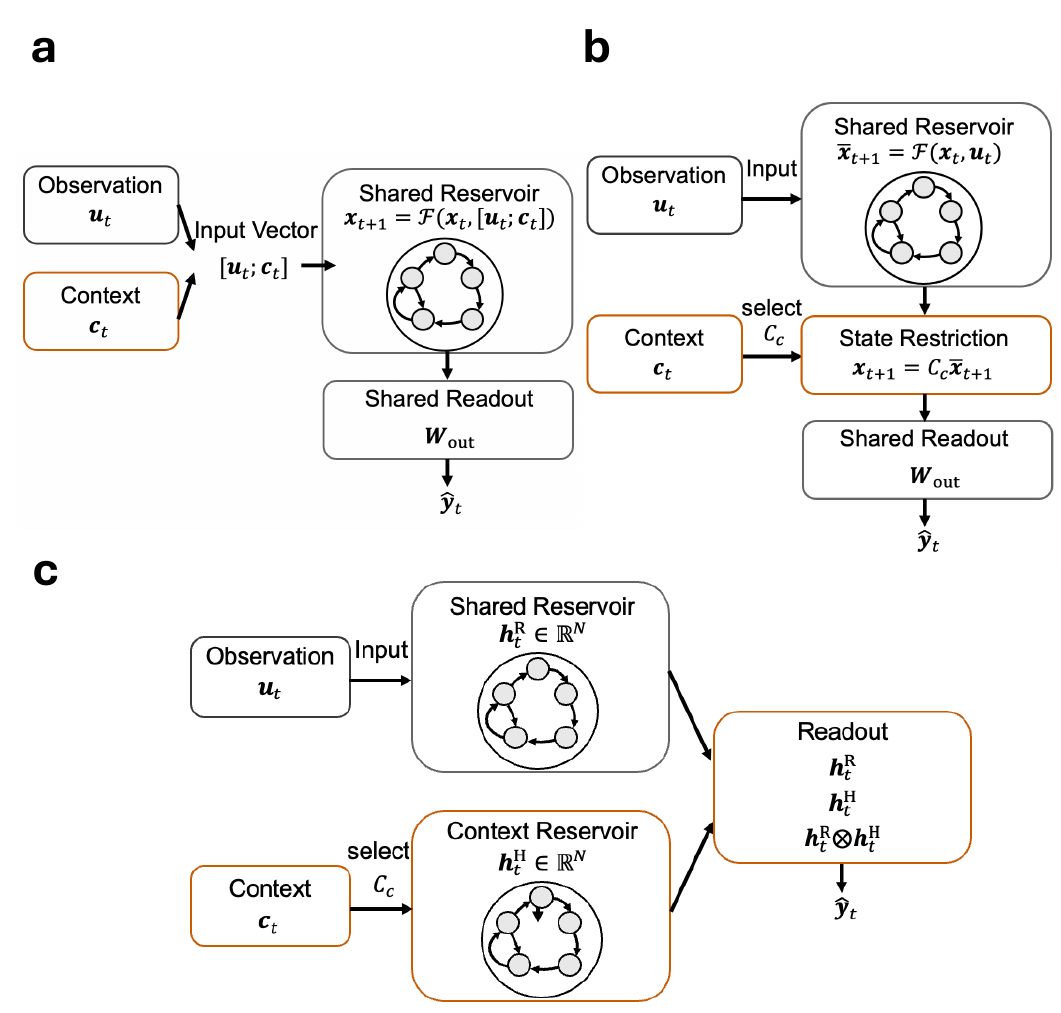}
    \caption{
    Three mechanisms for introducing contextual information into a shared reservoir system.
    (a) In the context-input ESN, context enters through the external input while the recurrent operator and readout remain shared.
    (b) In the full-matrix Conceptor model, context selects a state-space operator $C_k$ that filters the provisional reservoir state.
    (c) In the HyperReservoir, an additional context reservoir encodes the supplied context, and the bilinear term $\mathbf{h}^{\mathrm{H}}_{t}\otimes\mathbf{h}^{\mathrm{R}}_{t}$ allows this context representation to modulate the coefficients applied to the shared main-reservoir features.
    The recurrent matrices remain fixed in all three architectures.
    }
    \label{fig:architectures}
\end{figure*}

We now define the three contextual reservoir architectures compared in this study.
They receive the same observed variables and explicit contextual information, but differ in where that context acts on the computation.
Figure~\ref{fig:architectures} summarizes the context-input ESN, the full-matrix Conceptor, and the proposed HyperReservoir.
In the conventional context-input echo state network (ESN), context modifies the external input to the reservoir.
In the Conceptor model, context selects a state-space operator that acts on the reservoir state.
In the HyperReservoir, context instead modifies the readout applied to a shared reservoir representation.

\subsection{Context-input echo state network}
\label{subsec:context_input_esn}
We first consider a conventional ESN.
Its reservoir state $\mathbf{x}_{t}\in\mathbb{R}^{N}$ evolves according to
\begin{equation}
    \mathbf{x}_{t+1}
    =
    \mathbf{x}_{t}
    +
    \alpha
    \left[
        -\mathbf{x}_{t}
        +
        J\phi(\mathbf{x}_{t})
        +
        W_{\mathrm{in}}\mathbf{u}_{t}
    \right],
    \label{eq:esn_update}
\end{equation}
and the prediction is obtained from the linear readout,
\begin{equation}
    \widehat{\mathbf{y}}_{t}
    =
    W_{\mathrm{out}}\mathbf{x}_{t}
    +
    \mathbf{b}_{\mathrm{out}}.
    \label{eq:esn_output}
\end{equation}
Here, $J\in\mathbb{R}^{N\times N}$ is the recurrent matrix, $W_{\mathrm{in}}$ is the input matrix, $\alpha$ is the leak coefficient, and $\phi(\cdot)$ is an element-wise nonlinearity.
The recurrent and input matrices are randomly initialized and remain fixed, whereas the output coefficients are fitted by ridge regression.

For the context-input baseline, $\mathbf{u}_{t}$ denotes the input available to the reservoir, including both the observed dynamical state and the explicit context signal.
Thus, context affects prediction through the reservoir state, while the recurrent matrix and output transformation remain shared across contexts (see Fig.~\ref{fig:architectures}a).

\subsection{Conceptor-based state modulation}
\label{subsec:conceptor}
A different strategy is to introduce context directly at the level of the reservoir state.
Conceptors associate each stored pattern or context with a linear operator constructed from the corresponding reservoir-state distribution~\cite{jaeger2017using}.

In this study, we use the full-matrix form of the Conceptor, so that each context is represented by a matrix $\mathbf{C}_k \in \mathbb{R}^{N\times N}$ acting on the complete reservoir state.
Unlike a diagonal state-wise scaling, the full-matrix operator can filter arbitrary directions in reservoir state space, including directions defined by correlations among different reservoir units.

Let $\overline{\mathbf{x}}_{t+1}$ denote the provisional state obtained from the same fixed reservoir dynamics as in Eq.~\eqref{eq:esn_update}, with only the observed dynamical variables supplied to the reservoir.
For context $k$, the empirical state-correlation matrix is
\begin{equation}
    R_{k}
    =
    \left\langle
        \overline{\mathbf{x}}_{t}
        \overline{\mathbf{x}}_{t}^{\mathsf T}
    \right\rangle_{t\in\mathcal{T}_{k}},
    \label{eq:conceptor_correlation}
\end{equation}
where $\mathcal{T}_{k}$ contains the training samples associated with that context.
The corresponding full-matrix Conceptor is
\begin{equation}
    C_{k}
    =
    R_{k}
    \left(
        R_{k}
        +
        \gamma^{-2}I
    \right)^{-1},
    \label{eq:conceptor_definition}
\end{equation}
where $\gamma>0$ is the aperture parameter.

During prediction, the context selects the corresponding Conceptor and the provisional reservoir state is filtered according to
\begin{equation}
    \mathbf{x}_{t+1}
    =
    C_{k}\overline{\mathbf{x}}_{t+1}.
    \label{eq:conceptor_filtered_update}
\end{equation}
The output is then obtained from the shared readout
\begin{equation}
    \widehat{\mathbf{y}}_{t}
    =
    W_{\mathrm{out}}\mathbf{x}_{t}
    +
    \mathbf{b}_{\mathrm{out}}.
    \label{eq:conceptor_output}
\end{equation}

The eigenvalues of $C_k$ lie between zero and one, so the operator acts as a soft state-space filter that retains directions strongly represented by the training states while attenuating weakly represented directions.
The aperture $\gamma$ controls the strength of this restriction.
Each $C_k$ is estimated only from training reservoir states, and a single output matrix is fitted to the Conceptor-filtered states.
Thus, contextual dependence enters through the state representation while the readout remains shared (see Fig.~\ref{fig:architectures}b).

\subsection{Proposed HyperReservoir for context-dependent readout modulation}
\label{subsec:hyper_esn}
We next consider readout-level contextual adaptation.
Rather than modifying the main-reservoir state, the HyperReservoir keeps the main recurrent dynamics fixed and allows context to change how its representation is decoded.
Related reservoir-computing approaches have used multiple readout modules~\cite{laan2017multiple,tanaka2023multiple}, generalized nonlinear readouts~\cite{ohkubo2024generalized}, and separate reservoirs for measurement and regime-parameter sequences~\cite{zhong2017double}.
The HyperReservoir uses a specific combination of these ideas: observation and context are processed separately, and their interaction enters through a bilinear readout.

The HyperReservoir contains two fixed recurrent subsystems.
The main reservoir, with state $\mathbf{h}^{\mathrm{R}}_{t}\in\mathbb{R}^{N}$, is driven by the observed dynamical state:
\begin{equation}
    \mathbf{h}^{\mathrm{R}}_{t+1}
    =
    \mathbf{h}^{\mathrm{R}}_{t}
    +
    \alpha_{\mathrm{R}}
    \left[
        -\mathbf{h}^{\mathrm{R}}_{t}
        +
        J^{\mathrm{R}}
        \phi(\mathbf{h}^{\mathrm{R}}_{t})
        +W_{\mathrm{in}}^{\mathrm{R}}\mathbf{u}_{t}
    \right].
    \label{eq:hyper_main_reservoir}
\end{equation}
The context reservoir, with state $\mathbf{h}^{\mathrm{H}}_{t}\in\mathbb{R}^{M}$, is driven only by the explicit contextual signal $\mathbf{c}_{t}$:
\begin{equation}
    \mathbf{h}^{\mathrm{H}}_{t+1}
    =
    \mathbf{h}^{\mathrm{H}}_{t}
    +
    \alpha_{\mathrm{H}}
    \left[
        -\mathbf{h}^{\mathrm{H}}_{t}
        +
        J^{\mathrm{H}}
        \phi(\mathbf{h}^{\mathrm{H}}_{t})
        +
        W_{\mathrm{in}}^{\mathrm{H}}\mathbf{c}_{t}
    \right].
    \label{eq:hyper_context_reservoir}
\end{equation}
The recurrent matrices $J^{\mathrm{R}}$ and $J^{\mathrm{H}}$ and their input matrices remain fixed after initialization.
The main reservoir provides nonlinear features of the observed dynamics, whereas the smaller context reservoir provides a low-dimensional representation of the supplied context (see Fig.~\ref{fig:architectures}c).

Separate reservoir pathways for observed variables and regime information have previously been used for multi-regime time-series prediction~\cite{zhong2017double}.
In the present experiments, the supplied context is constant within each trial.
We therefore do not assume that recurrent memory in the context pathway is itself essential; the question addressed here is how the resulting context representation acts on the shared computation.

The principal construction considered here is the augmented HyperReservoir, whose readout feature vector is
\begin{equation}
    \boldsymbol{\psi}^{\mathrm{aug}}_{t}
    =
    \begin{bmatrix}
        \mathbf{h}^{\mathrm{R}}_{t} \\
        \mathbf{h}^{\mathrm{H}}_{t} \\
        \mathbf{h}^{\mathrm{H}}_{t}
        \otimes
        \mathbf{h}^{\mathrm{R}}_{t}
    \end{bmatrix},
    \label{eq:hyper_augmented_features}
\end{equation}
where $\otimes$ denotes the Kronecker product.
The corresponding prediction is
\begin{equation}
    \widehat{\mathbf{y}}_{t}
    =
    W^{\mathrm{R}}\mathbf{h}^{\mathrm{R}}_{t}
    +
    W^{\mathrm{H}}\mathbf{h}^{\mathrm{H}}_{t}
    +
    W^{\mathrm{RH}}
    \left(
        \mathbf{h}^{\mathrm{H}}_{t}
        \otimes
        \mathbf{h}^{\mathrm{R}}_{t}
    \right)
    +
    \mathbf{b}_{\mathrm{out}}.
    \label{eq:hyper_augmented_output}
\end{equation}
All coefficients in this readout are fitted by ridge regression; neither recurrent subsystem is trained by backpropagation.

To expose the context-dependent structure of the readout, we partition $W^{\mathrm{RH}}$ into $M$ blocks,
\begin{equation}
    W^{\mathrm{RH}}
    =
    \begin{bmatrix}
        B_{1} &
        B_{2} &
        \cdots &
        B_{M}
    \end{bmatrix},
    \label{eq:hyper_block_decomposition}
\end{equation}
where $B_m\in\mathbb{R}^{D_{\mathrm{out}}\times N}$.
Equation~\eqref{eq:hyper_augmented_output} can then be written as
\begin{equation}
    \widehat{\mathbf{y}}_{t}
    =
    \left[
        W^{\mathrm{R}}
        +
        \sum_{m=1}^{M}
        h^{\mathrm{H}}_{t,m}B_m
    \right]
    \mathbf{h}^{\mathrm{R}}_{t}
    +
    W^{\mathrm{H}}\mathbf{h}^{\mathrm{H}}_{t}
    +
    \mathbf{b}_{\mathrm{out}}.
    \label{eq:hyper_effective_readout_output}
\end{equation}
The effective coefficient matrix applied to the main-reservoir features is
\begin{equation}
    W_{\mathrm{eff}}
    \left(
        \mathbf{h}^{\mathrm{H}}_{t}
    \right)
    =
    W^{\mathrm{R}}
    +
    \sum_{m=1}^{M}
        h^{\mathrm{H}}_{t,m}B_m.
    \label{eq:hyper_effective_readout}
\end{equation}

Equation~\eqref{eq:hyper_effective_readout} provides the central interpretation of the architecture.
The matrix $W^{\mathrm{R}}$ defines a context-independent baseline mapping, while the remaining term provides a context-dependent correction.
The effective readout therefore forms an affine family parameterized by the context state,
so that shared predictive structure can remain in $W^{\mathrm{R}}$ while only the context-dependent part of the mapping changes.

This parameterization is closely related to the Hypernetwork viewpoint, in which one network or representation determines parameters used by another computation~\cite{ha2017hypernetworks}.
Here, however, the context dependence is restricted to the readout, and all fitted coefficients are obtained in a single ridge-regression problem.

This construction is related to previous reservoir architectures with multiple readouts~\cite{laan2017multiple,tanaka2023multiple}, but does not assign an independently fitted output matrix to each regime.
Instead, the effective readouts are coupled through the shared baseline $W^{\mathrm{R}}$, the matrices $\{B_m\}_{m=1}^{M}$, and the low-dimensional context representation $\mathbf{h}^{\mathrm{H}}_{t}$.

The bilinear feature block is also related to generalized nonlinear reservoir readouts~\cite{ohkubo2024generalized}.
Here, however, the multiplicative terms are specifically cross-interactions between the separately driven context and main-reservoir states.
This structure gives the bilinear term the interpretation of a context-dependent correction to the main-reservoir readout rather than a generic nonlinear feature expansion.

Alternative additive and multiplicative readout constructions are compared in Sec.~\ref{sec:readout_structure}.

%% file: sections/03_numerical_experiments.tex
\section{Numerical experiments}
\label{sec:numerical_experiments}
\begin{figure*}[t]
    \centering
    \includegraphics[width=\textwidth]{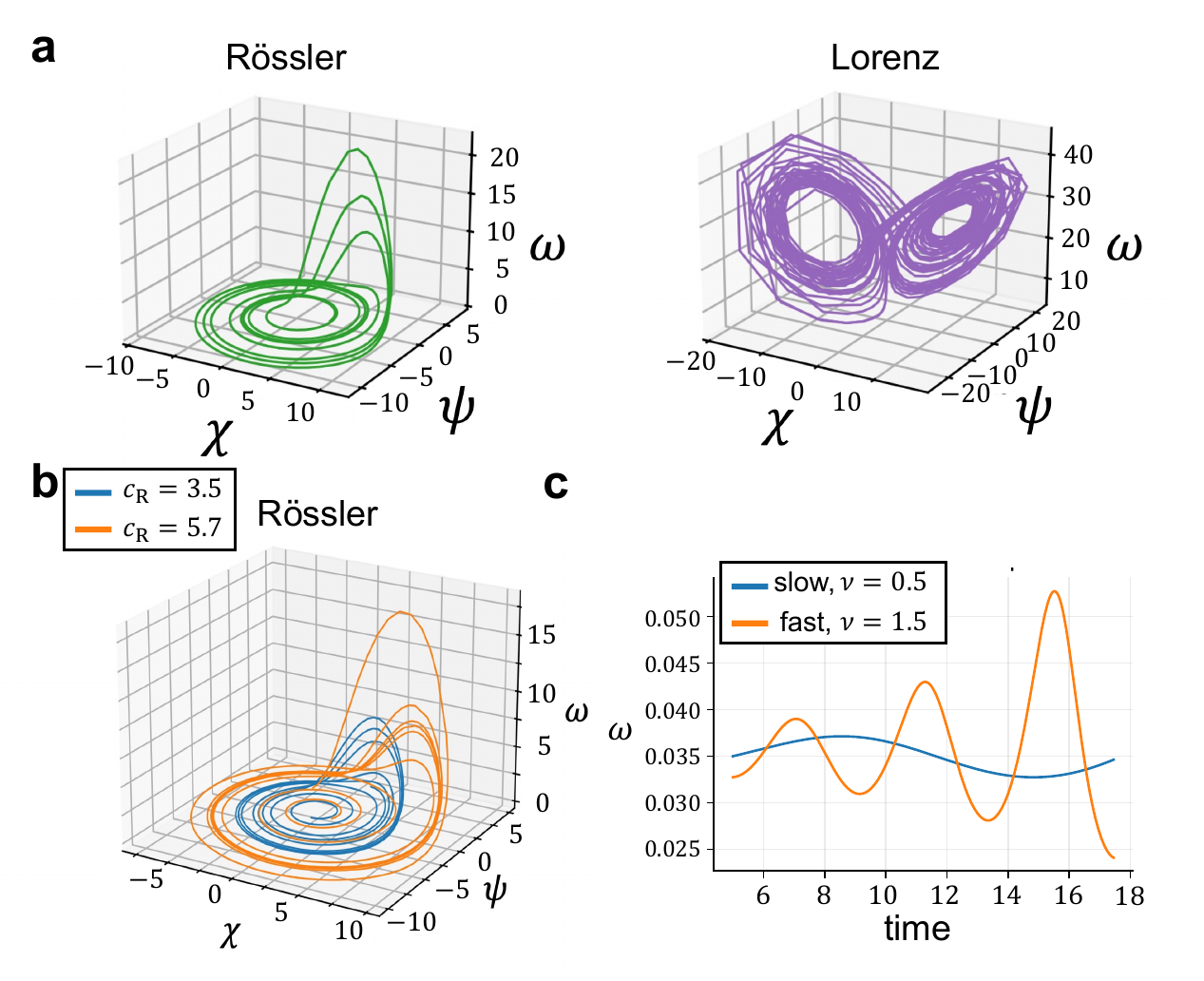}
    \caption{
    Dynamical settings used for contextual future-state prediction.
    (a) Geometrically distinct R\"ossler and Lorenz attractors.
    (b) Two R\"ossler regimes with $c_R=3.5$ and $c_R=5.7$.
    (c) The same R\"ossler attractor traversed at two temporal scales,
    $\nu_{\mathrm{slow}}=0.5$ and $\nu_{\mathrm{fast}}=1.5$.
    Scaling the complete vector field preserves the continuous-time orbit geometry while changing the rate of traversal.
    Representative traces of the $\omega$ component illustrate the resulting temporal difference.
    }
    \label{fig:task_progression}
\end{figure*}

We compare the three architectures on three future-state prediction settings based on Lorenz and R\"ossler dynamics, summarized in Fig.~\ref{fig:task_progression}.
The settings range from geometrically distinct attractors to a shared attractor traversed at different temporal scales.

In all experiments, the context is explicitly supplied, remains fixed within each trial, and is available at every time step.
All architectures use the same trajectories, data partitions, prediction targets, and evaluation metric.

\subsection{Geometrically distinct Lorenz and R\"ossler dynamics}
\label{subsubsec:lorenz_rossler}

The first task combines the Lorenz and R\"ossler systems.
The Lorenz dynamics are
\begin{align}
    \dot{\chi} &= \sigma(\psi-\chi),
    \label{eq:lorenz_x}
    \\
    \dot{\psi} &= \chi(\rho-\omega)-\psi,
    \label{eq:lorenz_y}
    \\
    \dot{\omega} &= \chi \psi-\beta \omega,
    \label{eq:lorenz_z}
\end{align}
with
\begin{equation}
    \sigma=10,
    \qquad
    \rho=28,
    \qquad
    \beta=\frac{8}{3}.
    \label{eq:lorenz_parameters}
\end{equation}

The R\"ossler dynamics are
\begin{align}
    \dot{\chi} &= -\psi-\omega,
    \label{eq:rossler_x}
    \\
    \dot{\psi} &= \chi+a_{\mathrm{R}}\psi,
    \label{eq:rossler_y}
    \\
    \dot{\omega} &= b_{\mathrm{R}}+\omega(\chi-c_{\mathrm{R}}),
    \label{eq:rossler_z}
\end{align}
with
\begin{equation}
    a_{\mathrm{R}}=0.2,
    \qquad
    b_{\mathrm{R}}=0.2,
    \qquad
    c_{\mathrm{R}}=5.7.
    \label{eq:rossler_step1_parameters}
\end{equation}

Each trial contains a trajectory from one of the two systems, whose identity is supplied as a two-dimensional one-hot context vector.
The two systems have clearly distinct attractor geometries.

\subsection{Related R\"ossler regimes}
\label{subsubsec:rossler_family}

The second task considers two regimes belonging to the same R\"ossler family.
The vector field is again given by Eqs.~\eqref{eq:rossler_x}--\eqref{eq:rossler_z}, with $a_{\mathrm{R}}=b_{\mathrm{R}}=0.2$, while
\begin{equation}
    c_{\mathrm{R}}\in\{3.5,\,5.7\}.
    \label{eq:rossler_context_parameters}
\end{equation}

The two values are represented by a two-dimensional one-hot context vector and are both included in the training, validation, and test sets.
Thus, the experiment concerns prediction for known regimes rather than interpolation to an unseen value of $c_{\mathrm{R}}$.

\subsection{Same attractor under different temporal scales}
\label{subsubsec:same_attractor}
The third setting uses the same R\"ossler system in both contexts but changes its temporal scale.
Both regimes use the R\"ossler system with $a_{\mathrm{R}}=b_{\mathrm{R}}=0.2$ and $c_{\mathrm{R}}=5.7$, but the complete vector field is multiplied by a context-dependent temporal-scale factor:
\begin{equation}
    \dot{\mathbf{\phi}}
    =
    \nu_k\mathbf{f}_{\mathrm{R}}(\mathbf{\phi}),
    \qquad
    \mathbf{\phi}
    =
    \begin{bmatrix}
        \chi & \psi & \omega
    \end{bmatrix}^{\mathsf T}.
    \label{eq:rossler_temporal_scale}
\end{equation}

The two temporal regimes are
\begin{equation}
    \nu_{\mathrm{slow}}=0.5,
    \qquad
    \nu_{\mathrm{fast}}=1.5.
    \label{eq:temporal_scale_values}
\end{equation}
The supplied context is $(c_{\mathrm{R}},\nu_k)$.
Since $c_{\mathrm{R}}=5.7$ in both regimes, only $\nu_k$ distinguishes the slow and fast cases.
For constant $\nu_k$, scaling the complete vector field amounts to a reparameterization of time: the continuous-time orbit geometry is unchanged, but the trajectory is traversed at a different rate.
Consequently, the same current state can require different future-state predictions over a fixed prediction interval.

\subsection{Prediction protocol}
\label{subsec:prediction_protocol}
\label{subsec:data_generation}

All dynamical systems are integrated using a fourth-order Runge--Kutta method with an internal integration step of $0.005$ and are sampled every $0.05$ continuous-time units.
Each trajectory contains $1000$ recorded samples.
The training, validation, and test sets contain $256$, $64$, and $64$ independently initialized trajectories, respectively.
The same generated datasets are used for all architectures.
Each architecture is evaluated using three independently initialized reservoir realizations, corresponding to seeds $s\in\{0,1,2\}$.
Additional integration settings, initial-condition distributions, and reproducibility details are provided in the Supplemental Material.

The task is one-step-ahead future-state prediction.
At time $t$, the observed state is
\begin{equation}
    \mathbf{s}_{t}
    =
    \begin{bmatrix}
        \chi_t & \psi_t & \omega_t
    \end{bmatrix}^{\mathsf T},
\end{equation}
and the supervised target is
\begin{equation}
    \mathbf{y}_{t}
    =
    \mathbf{s}_{t+H},
    \qquad
    H=1.
    \label{eq:prediction_target}
\end{equation}
Because the sampling interval is $0.05$, the corresponding physical prediction horizon is
\begin{equation}
    T_{\mathrm{pred}}
    =
    H\Delta t_{\mathrm{samp}}
    =
    0.05.
    \label{eq:physical_prediction_horizon}
\end{equation}

The first $20$ samples of each trajectory are discarded as reservoir washout, and samples without a valid future target are excluded from both fitting and evaluation.
During evaluation, the observed dynamical state is supplied to the reservoir at every time step; predicted states are not fed back recursively.
The reported errors therefore measure one-step prediction along observed trajectories rather than autonomous attractor generation.

All normalization statistics are estimated from the training set and then applied unchanged to the validation and test sets.
A single set of component-wise statistics is shared across contextual regimes within each task; in particular, the Lorenz and R\"ossler trajectories are not standardized separately.
This avoids introducing regime identity through preprocessing.
Further details are provided in the Supplemental Material.

\subsection{Recurrent-state allocation and parameter selection}
\label{subsec:context_routing}

The same observations and contextual information are available to all architectures, but the context is routed to different computational stages as defined in Sec.~\ref{sec:contextual_reservoir}.
Because the HyperReservoir contains an additional context reservoir, we control the total number of recurrent state variables across architectures.
We fix this total dimension to
\begin{equation}
    N_{\mathrm{tot}}=120.
    \label{eq:total_state_budget}
\end{equation}
The conventional ESN and full-Conceptor model therefore use $120$ recurrent states in their reservoir.
For the HyperReservoir, the same budget is divided between the main and context reservoirs:
\begin{equation}
    N+M=120.
    \label{eq:hyper_state_budget}
\end{equation}
The context-reservoir dimension is evaluated over
\begin{equation}
    M\in\{2,5,10,20,40\},
    \qquad
    N=120-M.
    \label{eq:M_sweep}
\end{equation}

For each dynamical setting, one value of $M$ is selected using the mean validation cNMSE over the three reservoir realizations and is then used for all test evaluations.
This procedure gives
\begin{equation}
    (N,M)
    =
    \begin{cases}
        (110,10), & \text{Lorenz--R\"ossler},\\
        (115,5), & \text{related R\"ossler},\\
        (115,5), & \text{same attractor}.
    \end{cases}
    \label{eq:selected_hyper_dimensions}
\end{equation}

The Conceptor aperture is selected in the same manner, using only validation data.
The resulting values are
\begin{equation}
    \gamma^\ast
    =
    \begin{cases}
        4, & \text{Lorenz--R\"ossler},\\
        1, & \text{related R\"ossler},\\
        1, & \text{same attractor}.
    \end{cases}
\end{equation}
The tested aperture values and other selection details are given in the Supplemental Material.

Both $M$ and $\gamma$ are selected from validation data using the mean over the three reservoir realizations; test data are not used for parameter selection.

The ridge coefficient is fixed to
\begin{equation}
    \lambda = 10^{-4} .
    \label{eq:ridge_coefficient}
\end{equation}

The recurrent spectral radii, input scales, and leak coefficients are listed in the Supplemental Material.


Each Conceptor is constructed exclusively from post-washout training reservoir states; validation and test states are not used.

Fixing the number of recurrent state variables does not equalize the number of fitted coefficients, storage requirements, or arithmetic cost.
These differences are considered separately in Sec.~\ref{sec:readout_structure} and the Supplemental Material.

\subsection{Evaluation metric}
\label{subsec:evaluation_metric}
Prediction accuracy is quantified using component-normalized mean squared error (cNMSE).
For output component $d$, we define
\begin{equation}
    E_d
    =
    \frac{
        \sum_{b,t}
        m_{b,t}
        \left(
            \widehat{y}_{b,t,d}
            -
            y_{b,t,d}
        \right)^2
    }{
        \max
        \left[
            \sum_{b,t}
            m_{b,t}
            y_{b,t,d}^{2},
            \epsilon
        \right]
    },
    \label{eq:component_nmse}
\end{equation}
where $b$ indexes trajectories, $t$ indexes time, and $m_{b,t}$ is the evaluation mask.
Specifically, $m_{b,t}=1$ for post-washout samples with a valid future-state target and $m_{b,t}=0$ otherwise.
The constant $\epsilon=10^{-4}$ prevents numerical instability if the target energy in the denominator becomes very small.

The reported error is
\begin{equation}
    \mathrm{cNMSE}
    =
    \frac{1}{D}
    \sum_{d=1}^{D}E_d,
    \qquad
    D=3.
    \label{eq:cnmse}
\end{equation}
Lower cNMSE indicates more accurate prediction.

%% file: sections/04_multifunctional_prediction.tex
\subsection{Results}
\label{sec:multifunctional_prediction}

\begin{figure*}[t]
    \centering
    \includegraphics[width=\textwidth]{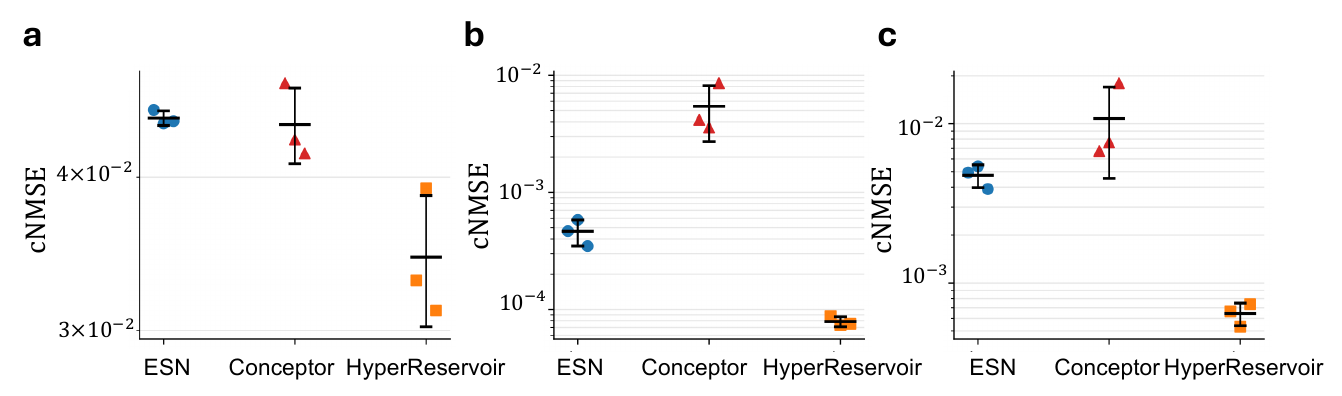}
    \caption{
    Future-state prediction for the three dynamical settings introduced in Sec.~\ref{sec:numerical_experiments}.
    (a) Geometrically distinct Lorenz and R\"ossler systems.
    (b) Two R\"ossler regimes with $c_{\mathrm{R}}=3.5$ and $c_{\mathrm{R}}=5.7$.
    (c) The same R\"ossler attractor traversed at different temporal scales,
    $\nu_{\mathrm{slow}}=0.5$ and $\nu_{\mathrm{fast}}=1.5$.
    Test component-normalized mean squared error (cNMSE) is shown for the
    context-input ESN, full-matrix Conceptor, and augmented HyperReservoir.
    Individual markers denote reservoir realizations; horizontal markers and
    error bars show the mean and sample standard deviation.
    The vertical axes are logarithmic, and lower cNMSE indicates more accurate prediction.
    }
    \label{fig:main_performance}
\end{figure*}

Fig.~\ref{fig:main_performance} shows the results of the numerical simulations across all three tasks. 
Figure~\ref{fig:main_performance}(a) shows the Lorenz--R\"ossler result.
All three architectures achieve mean cNMSE below $4.5\times10^{-2}$, with the ESN and Conceptor giving similar errors.
The augmented HyperReservoir gives the lowest error for each of the three reservoir realizations.

Figure~\ref{fig:main_performance}(b) shows the two R\"ossler regimes with $c_{\mathrm{R}}\in\{3.5,5.7\}$.
The separation among the architectures is substantially larger than in panel~(a).
The HyperReservoir gives the lowest error, while the Conceptor error is approximately one order of magnitude larger than that of the conventional ESN.

Figure~\ref{fig:main_performance}(c) shows the same-attractor setting with different temporal scales.
The HyperReservoir again gives the lowest test cNMSE, with a substantially larger separation from both reference models than in the Lorenz--R\"ossler setting. 

Taken together, the three tasks show that the augmented HyperReservoir provides a consistent advantage over the conventional context-input ESN and Conceptor.
The advantage is weaker on the first task, where the underlying target dynamics are geometrically distinct.
This pattern is consistent with the interpretation that context-dependent decoding becomes particularly useful when the representation implemented by the main reservoir can remain shared while the required predictive mapping changes across contexts.

Aggregate prediction errors do not show how these differences arise along individual trajectories.
We therefore examine the same-attractor setting in more detail in Sec.~\ref{sec:shared_attractor}.

%% file: sections/05_shared_attractor.tex
\section{Contextual separation on a shared attractor}
\label{sec:shared_attractor}
We examine the same-attractor setting using local prediction errors, context replacement, and a comparison of the corresponding Conceptor operators.
First, we compare the prediction errors under the two temporal regimes.
Second, we see how the prediction reacts to changing only the supplied contextual information while keeping the observed trajectory fixed.
Third, we compare the Conceptors learned for the two regimes to determine how strongly the temporal distinction is expressed in their reservoir-state geometry.

\begin{figure*}[t]
    \centering
    \includegraphics[width=\textwidth]{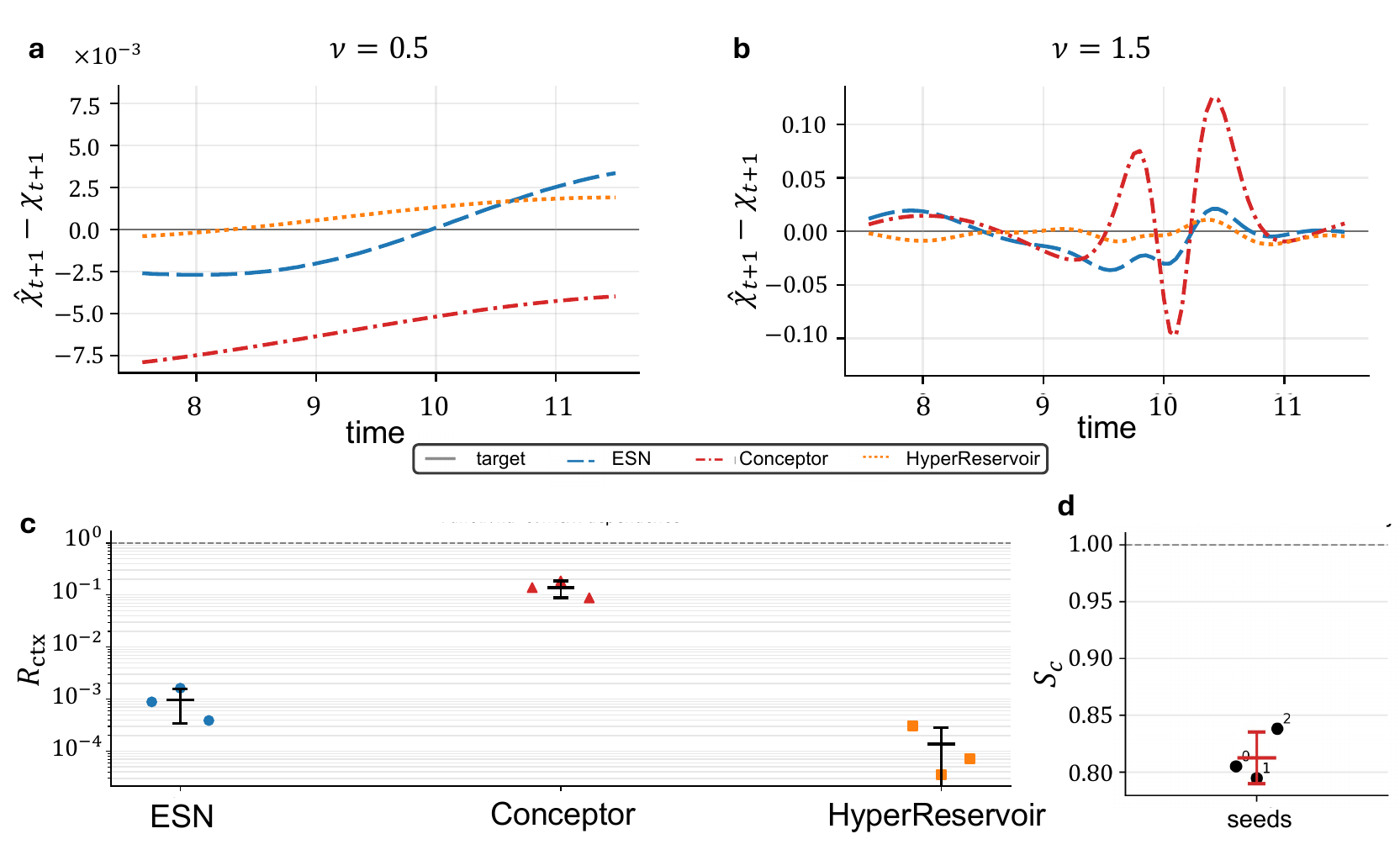}
    \caption{
    Analysis of contextual prediction on a shared attractor.
    (a,b) One-step prediction errors,
    $e_{\chi,t}=\hat{\chi}_{t+1}-\chi_{t+1}$,
    for the slow $\nu=0.5$ and fast $\nu=1.5$ regimes.
    The same representative reservoir realization and evaluation window are used for all models;
    the vertical scales differ between panels for visibility.
    (c) Dependence on the supplied context, quantified by the ratio
    $R_{\mathrm{ctx}}$ between correct- and wrong-context prediction errors.
    Lower values indicate a larger increase in error after context replacement.
    (d) Frobenius cosine similarity between the slow and fast Conceptors for the three reservoir realizations.
    Values close to unity indicate strong alignment of the two operators.
    }
    \label{fig:shared_attractor_mechanism}
\end{figure*}

\subsection{Finite-time prediction on a shared attractor}
\label{subsec:shared_attractor_prediction}

Figures~\ref{fig:shared_attractor_mechanism}(a) and
\ref{fig:shared_attractor_mechanism}(b) show the one-step prediction error of the first R\"ossler variable,
\begin{equation}
    e_{\chi,t}
    =
    \hat{\chi}_{t+1}-\chi_{t+1},
\end{equation}
for representative slow and fast test trajectories.
The HyperReservoir shows smaller local errors than the context-input ESN and full-matrix Conceptor in this realization, particularly in the fast regime, consistent with the aggregate errors in Fig.~\ref{fig:main_performance}.

\subsection{Correct- and wrong-context prediction}
\label{subsec:wrong_context}
To quantify how prediction depends on the supplied context, we replace the context while keeping the observed test trajectory fixed.

Let
\begin{equation}
    E_{ij}
    =
    \mathrm{cNMSE}
    \left(
        \text{true context } i,\,
        \text{supplied context } j
    \right)
    \label{eq:wrong_context_error}
\end{equation}
denote the prediction error under true context $i$ and supplied context $j$.
The diagonal terms $E_{11}$ and $E_{22}$ correspond to the correct context, whereas $E_{12}$ and $E_{21}$ correspond to context mismatch.

We summarize the dependence on context by the ratio
\begin{equation}
    R_{\mathrm{ctx}} = \frac{E_{11}+E_{22}} {E_{12}+E_{21}}. 
\end{equation}
A value close to unity indicates little sensitivity to context replacement, whereas a value substantially below unity indicates that prediction accuracy strongly depends on the supplied context.

For consistency with the main prediction comparison, the Conceptor wrong-context analysis uses the task-level aperture selected by mean validation cNMSE across the three reservoir realizations, $\gamma^\ast=1$.

Figure~\ref{fig:shared_attractor_mechanism}(c) shows that the HyperReservoir has the smallest
correct-to-wrong context error ratio.
Its prediction error therefore increases most strongly when the supplied context is replaced.
The conventional ESN shows an intermediate response, whereas the Conceptor ratio remains closer to unity.
Thus, the low prediction error of the HyperReservoir in this setting is accompanied by a strong dependence on the supplied context.

\subsection{Similarity of the slow and fast Conceptors}
\label{subsec:conceptor_similarity}
We next examine how strongly the slow and fast regimes are distinguished by the corresponding Conceptor operators.
The Conceptor construction represents each context through the second-order correlation structure of its reservoir states.
If the slow and fast temporal regimes induce strongly different reservoir-state geometries, their corresponding Conceptors should differ accordingly.
Therefore, we compare the operators learned for the two contexts.

We use the same validation-selected aperture, $\gamma^\ast=1$, as in the prediction comparison.

Let $C_{\mathrm{slow}}$ and $C_{\mathrm{fast}}$ denote the Conceptors obtained for the two temporal regimes.
Validation and test reservoir states are not used to estimate either Conceptor.

Their similarity is quantified using the Frobenius cosine
\begin{equation}
    S_C
    =
    \frac{
        \operatorname{tr}
        \left(
            C_{\mathrm{slow}}^{\mathsf T}
            C_{\mathrm{fast}}
        \right)
    }{
        \left\|C_{\mathrm{slow}}\right\|_F
        \left\|C_{\mathrm{fast}}\right\|_F
    }.
    \label{eq:conceptor_cosine}
\end{equation}
A value close to unity indicates strong alignment of the two operators in Frobenius space.

Figure~\ref{fig:shared_attractor_mechanism}(d) shows that the slow and fast Conceptors remain substantially aligned across reservoir realizations. 
Across the three reservoir realizations,
\begin{align}
S_C = 0.813 \pm 0.023, \label{eq:result_conceptor_cosine}
\end{align}
where the uncertainty denotes the sample standard deviation across the three reservoir realizations. 
A descriptive comparison of the ordered eigenvalue spectra is provided in Fig.~S2 of the Supplemental Material.

The two Conceptors are therefore not identical, but remain strongly aligned.
Together with the comparatively small effect of context replacement for the Conceptor model,
this indicates that the slow--fast distinction is only weakly expressed in the Conceptor-based state-space representation used here.

We next examine the structure of the HyperReservoir readout in Sec.~\ref{sec:readout_structure}.

%% file: sections/06_contextual_readout.tex
\section{Structure of the context-dependent readout}
\label{sec:readout_structure}
We compare three readout constructions to determine how the bilinear interaction contributes to the HyperReservoir prediction.
All three use the same main and context reservoirs and differ only in the features supplied to the readout.

The three feature constructions are
\begin{equation}
\begin{aligned}
    \boldsymbol{\psi}_{t}^{\mathrm{concat}}
    &=
    \begin{bmatrix}
        \mathbf{h}_{t}^{\mathrm{R}} \\
        \mathbf{h}_{t}^{\mathrm{H}}
    \end{bmatrix},
    \\
    \boldsymbol{\psi}_{t}^{\mathrm{strict}}
    &=
    \mathbf{h}_{t}^{\mathrm{H}}
    \otimes
    \mathbf{h}_{t}^{\mathrm{R}},
    \\
    \boldsymbol{\psi}_{t}^{\mathrm{aug}}
    &=
    \begin{bmatrix}
        \mathbf{h}_{t}^{\mathrm{R}} \\
        \mathbf{h}_{t}^{\mathrm{H}} \\
        \mathbf{h}_{t}^{\mathrm{H}}
        \otimes
        \mathbf{h}_{t}^{\mathrm{R}}
    \end{bmatrix}.
\end{aligned}
\label{eq:readout_variants}
\end{equation}

\begin{figure*}[t]
    \centering
    \includegraphics[width=\textwidth]
        {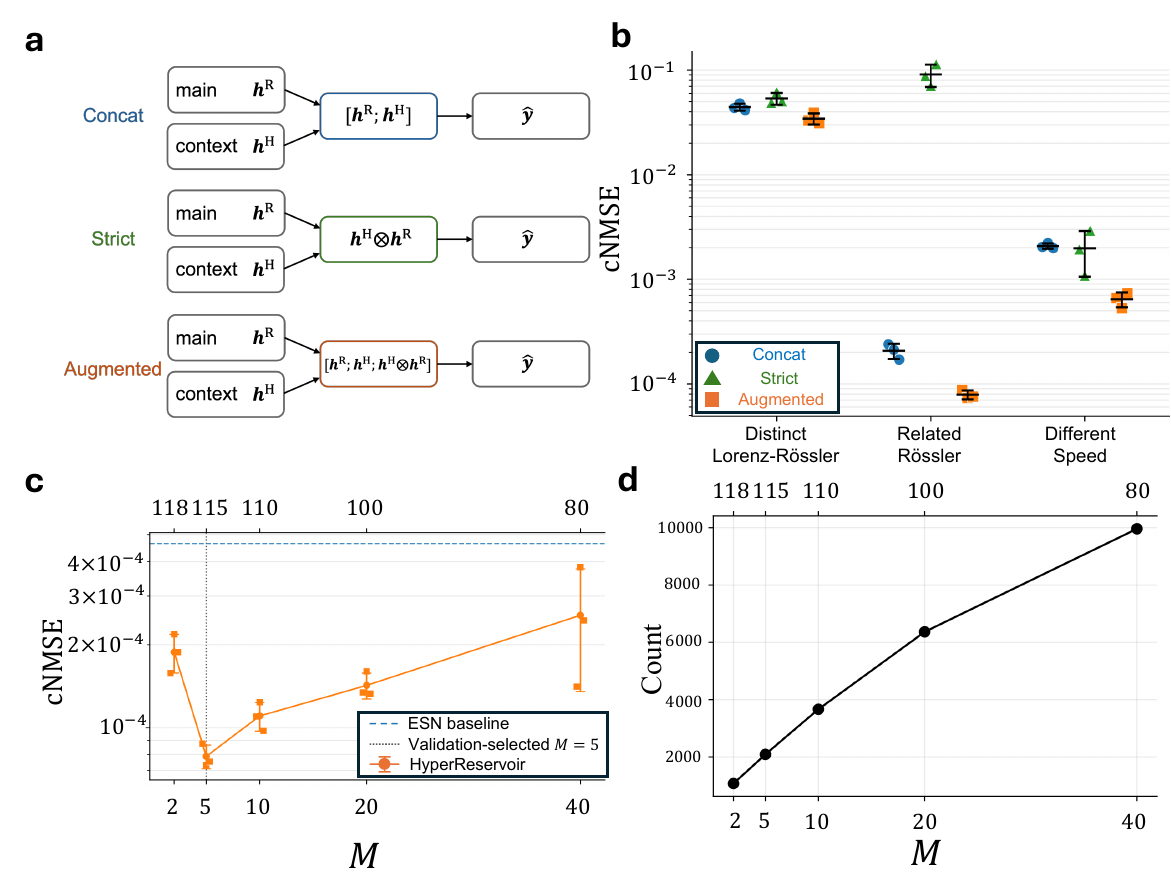}
    \caption{
    Structure and dimensionality of the HyperReservoir readout.
    (a) Three readout constructions.
    The Concat model uses the main- and context-reservoir states additively,
    the Strict model uses only their bilinear interaction, and the Augmented model combines both.
    (b) Test cNMSE for the three readout constructions across the dynamical settings introduced in Sec.~\ref{sec:numerical_experiments}.
    Individual markers denote reservoir realizations, and error bars show the mean and sample standard deviation.
    The Lorenz--R\"ossler setting uses \(N=110,M=10\), whereas the related-R\"ossler and same-attractor settings use \(N=115,M=5\).
    (c) Test cNMSE of the Augmented HyperReservoir for the related-R\"ossler setting as the context-reservoir dimension \(M\) is varied while \(N+M=120\).
    The dashed line denotes the corresponding context-input ESN result.
    (d) Number of fitted output coefficients as \(M\) is varied under the same constraint.
    }
    \label{fig:readout_ablation}
\end{figure*}

The \emph{Concat} readout provides direct additive access to both recurrent states but does not allow the coefficients applied to the main-reservoir features to depend on context.
The \emph{Strict} readout retains only the bilinear term and therefore implements a purely context-dependent multiplicative readout without an explicit context-independent main-reservoir mapping.
The \emph{Augmented} readout combines both additive features and the bilinear interaction and is the construction used in the principal comparisons.
They are visualized in Fig.~\ref{fig:readout_ablation}(a).

Figure~\ref{fig:readout_ablation}(b) shows that the Augmented readout gives the lowest mean test cNMSE in all three settings.
Both Concat and Strict give higher errors, indicating that neither additive access to the context state alone nor the bilinear interaction alone reproduces the performance of the complete Augmented readout.
The difference is particularly large for the related-R\"ossler setting.

The comparison also argues against readout dimension alone as an explanation for the performance difference.
Although the Augmented model has the largest readout, the Strict model already contains the \(NM\) bilinear feature block and therefore has a comparable number of fitted coefficients, yet its prediction error is substantially larger.
Detailed feature dimensions are reported in the Supplemental Material.

We next examine the dependence on the context-reservoir dimension $M$.
Keeping the total number of recurrent states fixed at $N+M=120$, we vary
\begin{equation}
    M\in\{2,5,10,20,40\},
    \label{eq:context_dimension_sweep}
\end{equation}
Increasing $M$ therefore reallocates recurrent states from the main reservoir to the contextual reservoir.

For the related-R\"ossler task shown in Fig.~\ref{fig:readout_ablation}(c), increasing $M$ does not produce a monotonic improvement in prediction.
The cNMSE first drops, and then starts to increase again.

The validation-selected dimensions were
\begin{equation}
    M^\ast
    =
    \begin{cases}
        10, & \text{Lorenz--R\"ossler},\\
        5, & \text{related R\"ossler},\\
        5, & \text{same attractor}.
    \end{cases}
    \label{eq:selected_context_dimensions}
\end{equation}
Thus, the validation procedure selects relatively small context reservoirs in all three settings.
Increasing the context-reservoir dimension beyond the validation-selected value does not systematically improve prediction and can instead reduce accuracy.

This behavior is consistent with the functional division between the two recurrent subsystems.
Because $N+M$ is fixed, increasing $M$ enlarges the context representation while reducing the number of states available to the main reservoir. 
The non-monotonic dependence is therefore consistent with a trade-off between these two state allocations.

Changing $M$ also changes the dimensionality of the Augmented readout.
Under $N+M=120$, the number of bilinear features is
\begin{equation}
    NM
    =
    M(120-M).
    \label{eq:bilinear_feature_count}
\end{equation}
Over the evaluated range, \(NM\) and the total number of fitted coefficients increase monotonically with \(M\), as shown in Fig.~\ref{fig:readout_ablation}(d).
Prediction accuracy does not improve monotonically over the same range.
Thus, the observed dependence on \(M\) cannot be explained simply by an increase in fitted readout dimension.

Together, the readout comparison and dimension sweep show that increasing context dimension or fitted feature count alone does not account for the prediction improvement of the Augmented HyperReservoir.

%% file: sections/07_conclusion.tex
\section{Conclusion}
\label{sec:conclusion}
We proposed the HyperReservoir, a reservoir-computing architecture for context-dependent decoding of shared recurrent dynamics.
A common main reservoir represents the observed dynamics, while a smaller context pathway modulates the readout applied to this representation.
In the augmented construction, the effective readout consists of a shared baseline mapping and a context-dependent correction.

We compared the HyperReservoir with two alternatives that introduce the same contextual information at different stages of the computation: a conventional context-input ESN and full-matrix Conceptor-based state modulation.
The three future-state prediction tasks ranged from geometrically distinct Lorenz and R\"ossler systems, through related R\"ossler regimes, to the same R\"ossler attractor traversed at different temporal scales.
The augmented HyperReservoir achieved the lowest mean test cNMSE in all three tasks.
We observed a modest advantage in the first task and larger advantages in the latter two, which is consistent with context-dependent decoding being most useful when more of the recurrent representation can remain shared while the required predictive mapping changes across contexts.

The temporal-scale task illustrates a form of multifunctionality complementary to the multistability studied in much of the reservoir-computing literature, where a single trained system supports multiple coexisting attractors~\cite{flynn2021multifunctionality,flynn2023seeing,du2025multifunctional}, including cases in which the corresponding trajectories overlap in state space~\cite{flynn2023seeing}.
The slow and fast regimes share the exact same continuous-time orbit geometry, but their finite-time flow maps differ.
Multifunctionality therefore need not require multiple attractors: different computations can correspond to different temporal evolutions on a common state-space structure.
The HyperReservoir realizes the corresponding separation between representation and interpretation: the main reservoir provides nonlinear features reused across regimes, and the context-dependent readout determines how those features contribute to the prediction.

Where context should enter a recurrent system depends on which part of the computation must vary across regimes.
In the context-input ESN, context changes the reservoir trajectory through external forcing, and a single readout must decode all contexts.
In the Conceptor model, context restricts the reservoir state space and therefore relies on regimes that induce distinguishable state distributions.
In the HyperReservoir, the representation is retained and context changes only its decoding.
For the geometrically distinct Lorenz and R\"ossler systems, state-level modulation remains effective: the Conceptor and the conventional ESN reach nearly identical errors.
In the same-attractor setting, the slow and fast Conceptors remain strongly aligned, and replacing the supplied Conceptor produces a comparatively smaller change in prediction than replacing the context in the HyperReservoir.
These observations are consistent with the temporal distinction being only weakly expressed in the Conceptor-based state-space representation used here.

The readout comparison shows that neither additive access to the context state nor the bilinear interaction alone reproduces the performance of the augmented construction.
The context-dimension sweep further shows that prediction does not improve monotonically with either context dimension or fitted readout size.

Several limitations define the scope of these conclusions.
During evaluation, the observed state is supplied at every time step, so the results concern short-horizon prediction rather than autonomous attractor generation.
Context is explicitly supplied, and inference of an unknown or switching context is not addressed.
The evaluated contexts are also a finite discrete set, so generalization to unseen regimes remains to be tested.
Finally, the comparison controls the total number of recurrent state variables, not the number of fitted coefficients, storage requirements, or arithmetic cost.

When substantial recurrent structure can remain shared while its predictive interpretation must change, context-dependent decoding provides a direct mechanism for contextual flexibility.
The HyperReservoir implements this principle while retaining training by linear regression.